\documentclass[11pt]{article}

\PassOptionsToPackage{dvipsnames}{xcolor}

\usepackage[final]{acl}

\usepackage{times}
\usepackage{latexsym}

\usepackage[T1]{fontenc}

\usepackage[utf8]{inputenc}

\usepackage{microtype}

\usepackage{inconsolata}

\usepackage{graphicx}
\usepackage{hyperref}
\usepackage{url}

\usepackage{enumitem}
\usepackage{amsmath}
\usepackage[utf8]{inputenc}
\usepackage[T1]{fontenc}
\usepackage{hyperref}
\usepackage{url}
\usepackage{booktabs}
\usepackage{amsfonts}
\usepackage{nicefrac}
\usepackage{microtype}
\usepackage{booktabs}
\usepackage{graphicx}
\usepackage{multirow}
\usepackage{subcaption}
\usepackage{listings}
\usepackage{lscape}
\usepackage{longtable}
\usepackage{placeins}
\usepackage{siunitx}
\usepackage{tcolorbox}
\usepackage{enumitem}
\usepackage{float}
\usepackage{xspace}
\usepackage{tabularx}
\usepackage{booktabs}
\usepackage{pgfplots}
\pgfplotsset{compat=1.17}
\usetikzlibrary{patterns}

\newcommand{\promptvar}[1]{\textit{<#1>}}
\tcbuselibrary{breakable}
\usepackage{colortbl}
\definecolor{mygreen}{HTML}{1E8449}
\definecolor{myred}{HTML}{C0392B}
\definecolor{mygray}{gray}{0.92}

\renewcommand{\arraystretch}{0.8}

\newtcolorbox{jsonexample}{
    colback=gray!10,
    colframe=gray!50,
    boxsep=4pt, left=8pt, right=4pt, top=4pt, bottom=4pt,
    sharp corners,
    breakable,
    before upper={\ttfamily}
}

\newtcolorbox{rqbox}{
    colback=gray!5,
    colframe=gray!60,
    leftrule=3pt,
    rightrule=0pt, toprule=0pt, bottomrule=0pt,
    boxsep=2pt, left=10pt, right=10pt, top=2pt, bottom=2pt,
    sharp corners,
    breakable,
    fontupper=\itshape
}

\DeclareRobustCommand{\method}{\textsc{GUI-PRA\xspace}}

\title{GUI-PRA: Process Reward Agents for GUI Tasks}

\author{
 \textbf{Tao Xiong}$^{1,\dagger,*}$,
 \textbf{Xavier Hu}$^{1,\dagger}$,
 \textbf{Yurun Chen}$^{1}$,
 \textbf{Yuhang Liu}$^{1}$,\\
 \textbf{Changqiao Wu}$^{2}$,
 \textbf{Pengzhi Gao}$^{2}$,
 \textbf{Wei Liu}$^{2}$,
 \textbf{Jian Luan}$^{2}$,
 \textbf{Shengyu Zhang}$^{1,\ddagger}$\\\\
 \vspace{2mm}
 $^{1}$Zhejiang University, $^{2}$MiLM Plus, Xiaomi Inc.\\
 \vspace{2mm}
 \texttt{\{xiongtao, sy\_zhang\}@zju.edu.cn}
}

\begin{document}
\maketitle

\renewcommand{\thefootnote}{}
\footnotetext{$^{\dag}$Equal contribution, $^{\ddag}$Corresponding Author, $^{*}$Work done during an internship at Xiaomi Inc.} 
\renewcommand{\thefootnote}{\arabic{footnote}}

\begin{abstract}

Long-horizon GUI automation remains challenging due to error accumulation over extended interaction sequences. Process Reward Models (PRMs) provide dense step-level supervision for mitigating error accumulation, yet standard PRMs are poorly suited to GUI verification. Standard PRM judgments often rely on superficial visual alignment rather than functional correctness, reflecting an \textit{evaluative knowledge gap} caused by missing domain-specific adjudication logic. Standard PRMs also perform passive, single-pass visual assessment, which creates \textit{Visual Ambiguity} when reliable judgment requires actively locating, parsing, or inspecting task-relevant UI evidence. We introduce \textbf{\method}, a Process Reward Agent that transforms GUI process evaluation from passive scoring into active investigation. \method~couples \textbf{Experience-Injected Criterion Synthesis}, which distills generalized verification principles into state-specific criteria, with \textbf{Criterion-Guided Autoregressive Perception}, which uses these criteria to navigate multi-granularity visual tools and gather grounded evidence. On AndroidWorld and Mobile-MiniWoB++, \method~achieves average gains of \textbf{5.0} and \textbf{6.5} SR points over standard PRMs across the Qwen-VL backbones, with Qwen3-VL attaining \textbf{54.74\%} success rate on AndroidWorld. On the offline OS-Critic Bench, \method~demonstrates strong competitiveness against fully trained critic models. We will release our codes at \url{https://github.com/YuanDaoze/GUI-PRA/}.

\end{abstract}

\section{Introduction}

\begin{figure}[!t]
    \centering
    \includegraphics[width=1.0\linewidth]{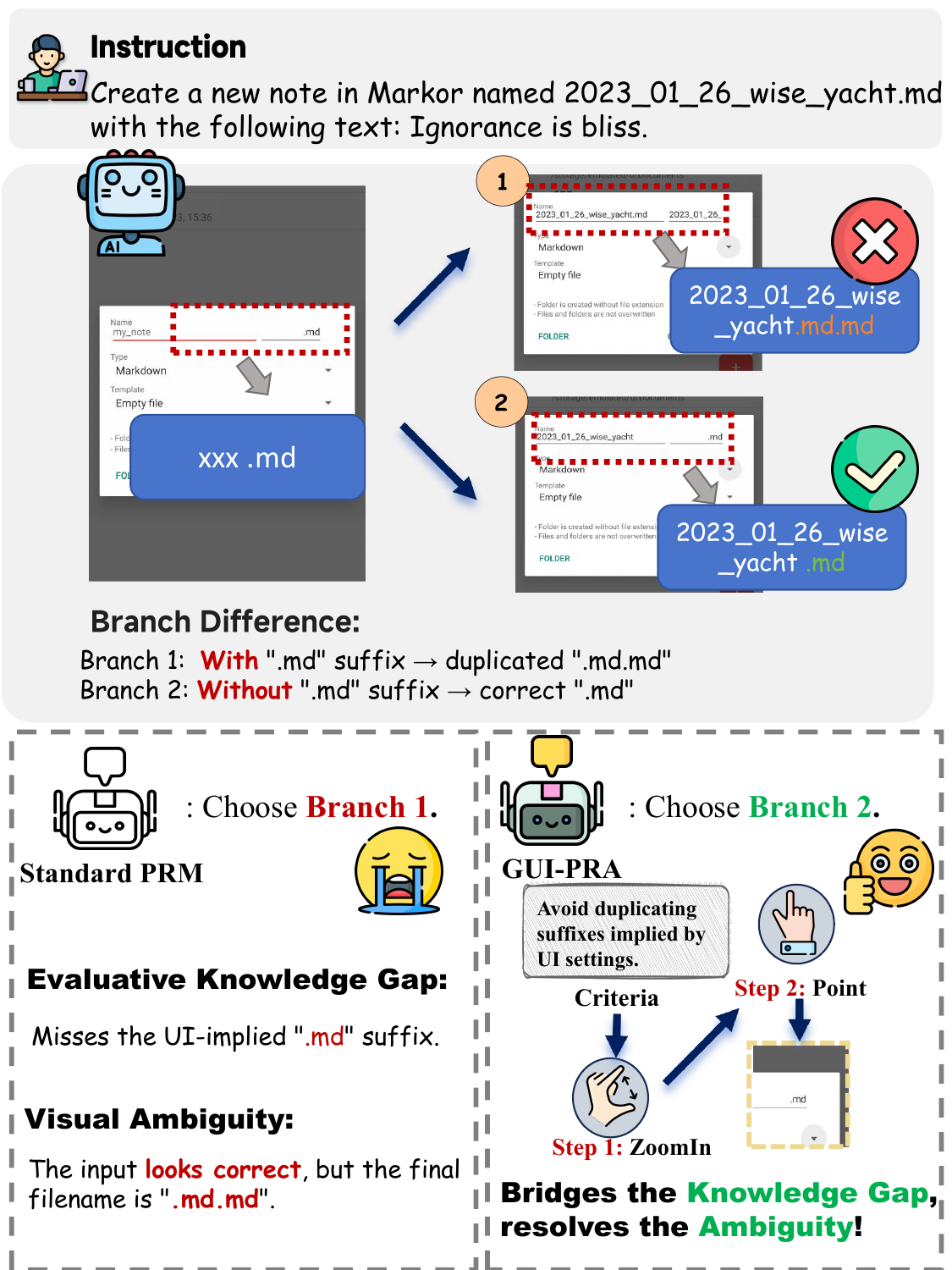}
    \caption{A motivating example of a ``Double Suffix'' error case, comparing Standard PRM and \method. In the figure, ``ZoomIn'' is shorthand for \texttt{ZoomInSubfigure}.}
    \label{fig:task}
\end{figure}

Graphical User Interface (GUI) Agents \citep{hu2025osagentssurveymllmbased,li2024personalllmagentsinsights,cao2026xiaomigui0technicalreport,ding2026uimateadvancingopenweightfoundation,wang2026deskcraftbenchmarkingdesktopagents}, powered by Multimodal Large Language Models (MLLMs) \citep{zhang2025instructiontuninglargelanguage,li2024ariaopenmultimodalnative, zheng2024minigpt5interleavedvisionandlanguagegeneration}, have shown promise for automating complex user tasks. Yet these agents frequently fail in long-horizon scenarios \citep{tian2025agentprogempoweringlonghorizongui, wang2025historyawarereasoningguiagents, wu2025autoscalingcontinuousmemorygui}, where small errors compound over extended interaction sequences.
GUI tasks particularly benefit from process supervision because feedback is sparse and a single incorrect action can invalidate the rest of a trajectory.
To mitigate this, Process Reward Models (PRMs) \citep{gandhi2025agentsastraycoursecorrectingswe, wanyan2025lookleapguicriticr1model, xiao2025uigenieselfimprovingapproachiteratively, chen2025guishepherdreliableprocessreward} have gained traction as a mechanism for providing dense, step-level supervision that guides agents toward correct trajectories and prunes erroneous paths during inference.

\begin{figure*}[!t]
    \centering
    \includegraphics[width=1\linewidth]{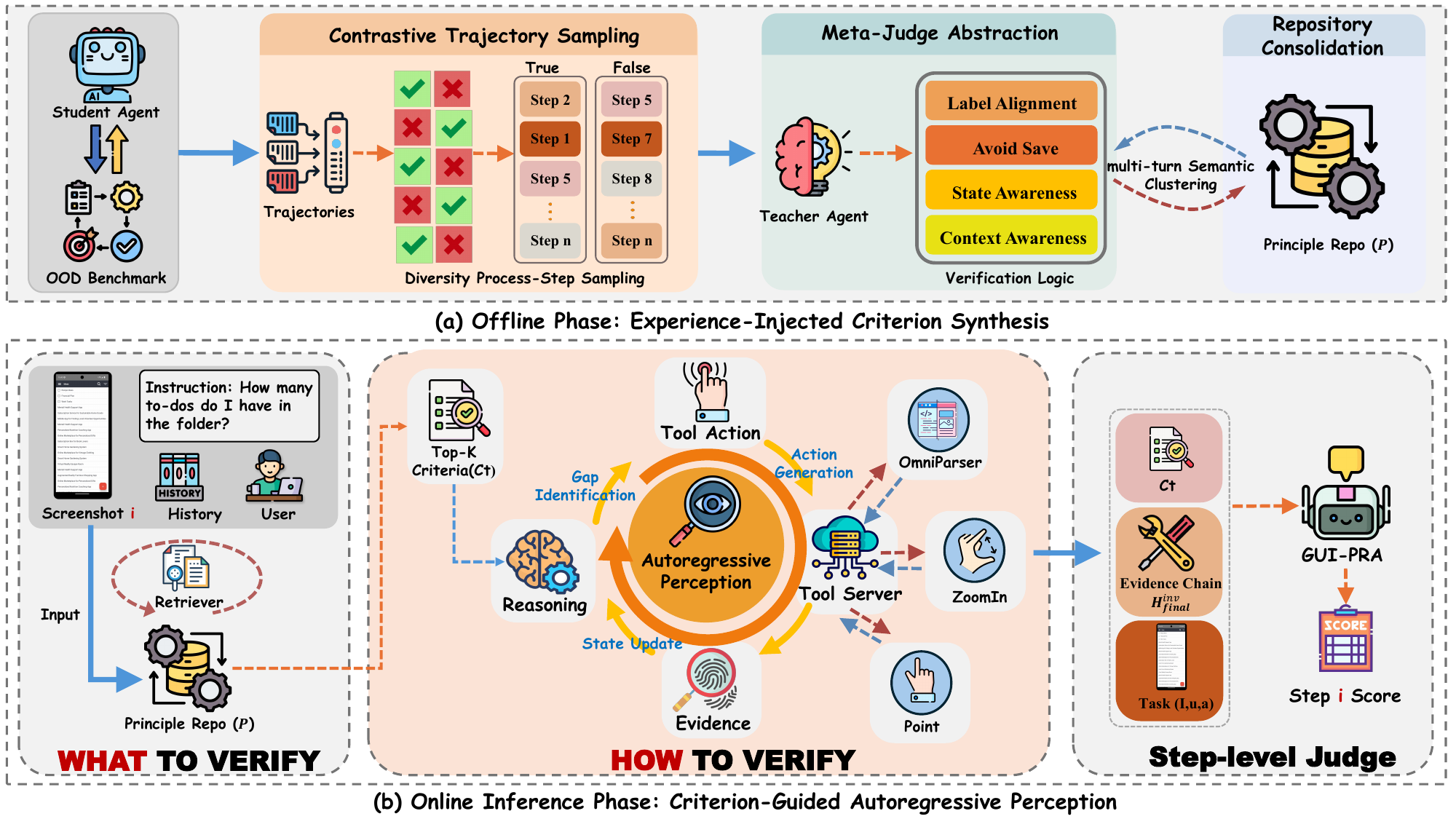}
    \caption{
    The overall framework of \method. 
    \textbf{Offline Phase:} We construct a Principle Repository ($\mathcal{P}$) by distilling generalized verification logic from contrastive trajectories using a Meta-Judge.
    \textbf{Online Phase:} The model synthesizes explicit criteria ($C_t$) based on $\mathcal{P}$, then enters a \textbf{Criterion-Guided Autoregressive Loop} to actively gather grounded visual evidence via tools (e.g., \texttt{ZoomInSubfigure}, \texttt{Point}) for the final evidence-based supervision.
}
    \label{fig:method}
\end{figure*}

However, directly applying PRMs in GUI domains yields suboptimal supervision due to two fundamental limitations. As illustrated in Figure~\ref{fig:task}, standard PRMs suffer from an \textit{evaluative knowledge gap} \citep{zhang2025lessonsdevelopingprocessreward, dai2026proreproactiverewardgui, krumdick2025freelabelslimitationsllmasajudge,chiu2025aideagenticallyimprovevisual}. Unlike domain experts who rely on deep functional knowledge to verify correctness, standard PRMs judge actions by superficial visual alignment. They lack the logic to verify procedural dependencies (e.g., confirming that preconditions are met before an action) or enforce state-transition rules (e.g., checking data persistence). This leads to failures such as the \textit{completion illusion}, where a PRM assigns high reward to an agent that has visually filled a form but failed to execute the functional ``Save'' action, or misses subtle violations of negative constraints.

Second, standard PRMs are constrained by \textit{Visual Ambiguity} \citep{fan2024readpointedlayoutawaregui,xu2024attentiondrivenguigroundingleveraging,hu2024visualprogramdistillationdistilling}. Standard PRM evaluation is typically passive and single-pass, leaving the judge unable to actively locate, parse, or inspect the UI evidence required for reliable assessment. Fine-grained cues such as checkbox states, dense text fields, or visually similar list items may therefore remain unresolved during scoring. Without an evidence-seeking mechanism, PRMs tend to render verdicts from coarse visual proxies rather than grounded observations, leading to hallucinatory conclusions.

We introduce \textbf{\method}, a Process Reward Agent that addresses both limitations through a single organizing principle: \textit{decoupling what to verify from how to verify it}. As shown in Figure~\ref{fig:method}, \method~transforms a standard PRM from a static scorer into an active investigator. To bridge the evaluative knowledge gap, we propose \textbf{Experience-Injected Criterion Synthesis}. Instead of relying on implicit intuition, this mechanism synthesizes explicit adjudication criteria for the current UI state by referencing a repository of verification principles distilled from generalized expert logic. These criteria establish a clear, rule-based standard for what constitutes success at each step. To overcome Visual Ambiguity, we propose \textbf{Criterion-Guided Autoregressive Perception}. Guided by the synthesized criteria, this mechanism transforms verification from passive viewing into an active evidence-seeking process. It uses multi-granularity visual tools, including global layout parsing and local inspection, to gather grounded evidence before rendering a verdict.

\begin{itemize}[itemsep=0pt, leftmargin=*]
    \item 
        We propose \textbf{\method}, a Process Reward Agent which shifts the supervision paradigm from static, global-view scoring to an active, evidence-seeking investigation process.
        
    \item 
        We introduce two coupled mechanisms: \textbf{Experience-Injected Criterion Synthesis} for deriving state-specific adjudication criteria, and \textbf{Criterion-Guided Autoregressive Perception} for gathering grounded visual evidence through tool use.
            
    \item 
        We conduct extensive experiments on AndroidWorld, Mobile-MiniWoB++, and OS-Critic Bench. \method~improves the Qwen-VL series by \textbf{5.0} and \textbf{6.5} SR points on average across the two online benchmarks, and remains competitive with trained critic models in offline evaluation.
\end{itemize}

\section{Related Work}

\paragraph{(M)LLM-based GUI Agents.}
Multimodal large language models \citep{bai2025qwen3vltechnicalreport, singh2026openaigpt5card, geminiteam2025geminifamilyhighlycapable} have substantially expanded the capability of GUI agents.
Recent work strengthens GUI agents by post-training MLLMs with supervised fine-tuning \citep{liu2025infiguir1advancingmultimodalgui, hong2024cogagentvisuallanguagemodel, wu2024osatlasfoundationactionmodel} or reinforcement learning \citep{liu2025infiguir1advancingmultimodalgui,xu2025mobilerlonlineagenticreinforcement,xu2026mobile,wang2026measuretwiceclickonce,wang2025rollingstonegathersmoss} on curated interaction trajectories.
A complementary line designs agent frameworks with specialized modules for planning, grounding, memory, and verification \citep{ye2025mobileagentv3fundamentalagentsgui, zhang2023appagentmultimodalagentssmartphone, xu2026mobile}, yet long-horizon GUI automation still suffers from step-level errors that can cascade into unsafe or irreversible outcomes~\citep{chen2025harmonyguardsafetyutilityweb, chen2025evaluatingrobustnessmultimodalagents}. This motivates process-level supervision, where an external verifier checks intermediate candidate actions to avoid unreliable trajectories.

\paragraph{Process Reward Models for GUI Agents.} Process-level supervision for GUI agents has emerged from three related sources. One line directly trains dedicated PRMs on GUI interaction data, using \citep{chen2025guishepherdreliableprocessreward, wanyan2025lookleapguicriticr1model, xiao2025uigenieselfimprovingapproachiteratively,wu2025osoraclecomprehensiveframeworkcrossplatform} to predict candidate-step validity. These methods require substantial domain-specific data and training resources, which can limit transferability. A second line repurposes general-purpose multimodal models as judges, either in an LLM-as-a-Judge setting over screenshots and trajectories or in an Agent-as-a-Judge setting over proposed thought-action pairs \citep{zheng2026adaptivemilestonerewardgui,dai2026proreproactiverewardgui,cui2026agenticrewardmodelingverifying}. Proactive and agentic approaches can gather additional evidence, but rely largely on free-form reasoning, leaving criteria implicit and evidence acquisition difficult to control. A third line uses PRM-style signals as stepwise feedback inside reinforcement learning, where principle-constrained, curriculum-staged, or retrospective rewards guide policy optimization rather than inference-time verification. Representative examples include \citet{nong2026craftguicurriculumreinforcedagentgui, lu2025orcuststepwisefeedbackreinforcementlearning,tang2025guig2gaussianrewardmodeling,feng2025groupingrouppolicyoptimizationllm}. These methods primarily optimize policy parameters rather than provide inference-time supervision for a frozen actor. We propose \method~, an evidence-grounded Process Reward Agent that guides GUI action selection at inference time using explicit criteria and active visual verification.

\section{Preliminary}
\label{sec:preliminary}

\subsection{GUI Task Automation}
We formulate GUI task automation as a sequential decision-making problem. Given a natural language instruction $I$ (formerly denoted as goal $g$) and an initial GUI state $S_0$, which comprises a screenshot $scr_0$ and structural elements $e_0$ (i.e., $S_0 = (scr_0, e_0)$), the agent must generate a sequence of actions $\hat{\alpha}$ to fulfill the instruction.
An action sequence $\hat{\alpha} = (a_1, a_2, \ldots, a_T)$ is valid if the terminal state satisfies a success predicate $V$, such that $V(\mathcal{T}(S_0, \hat{\alpha}), I) = \texttt{pass}$, where $\mathcal{T}$ denotes the environment transition function $S_{t+1} = \mathcal{T}(S_t, a_t)$.

The agent operates via a ReAct-style loop \citep{yao2023reactsynergizingreasoningacting}. At step $t$, the interaction history is denoted as:
\[
\mathcal{H}_t \;=\; \big(u_1,a_1,o_1,\,u_2,a_2,o_2,\,\ldots,\,u_t,a_t,o_t\big),
\]
where $u_i$ represents the agent's reasoning thought, $a_i$ is the executable action, and $o_i$ is the observation consisting of the new state $S_i$. The policy $\pi_\theta$ generates the next thought-action pair $(u_{t+1}, a_{t+1}) \sim \pi_\theta(\cdot \mid \mathcal{H}_t, I)$.

\subsection{Supervision with a Standard PRM}
\label{sec:proc-supervision}
To mitigate error accumulation in long-horizon tasks, a Process Reward Model (PRM) provides dense supervision. At step $t$, the policy generates a set of $k$ candidate thought-action pairs, denoted as $\mathcal{A}_t = \{(u_{t,j}, a_{t,j})\}_{j=1}^k$.
The standard PRM acts as a discriminator to select the optimal transition.
Formally, it estimates the probability of future success conditioned on the static history $\mathcal{H}_{t-1}$ and the current candidates:
\begin{equation}
\label{eq:prm}
(u_t,a_t) = \underset{(u, a) \in \mathcal{A}_t}{\text{argmax}} \;\; \text{PRM}(I, S_{t-1}, \mathcal{H}_{t-1}, u, a)
\end{equation}

\section{Methodology}
\label{sec:methodology}

\subsection{Overview}
As illustrated in Figure~\ref{fig:method}, \method~redefines the Process Reward Model from a passive scorer to an active investigator. To address two key limitations of standard PRMs, \method~adopts a three-phase pipeline: (1) \textbf{Experience-Injected Criterion Synthesis} (\S\ref{sec:criterion_synthesis}), which derives \textit{what} to verify from distilled verification principles; 
(2) \textbf{Criterion-Guided Autoregressive Perception} (\S\ref{sec:autoregressive_perception}), which uses active tools to determine \textit{how} to verify; 
and (3) \textbf{Evidence-Based Supervision} (\S\ref{sec:enhanced_supervision}), computing the final reward $r_t$ based on grounded evidence rather than static intuition.

\subsection{Experience-Injected Criterion Synthesis}
\label{sec:criterion_synthesis}

Experience-Injected Criterion Synthesis addresses the lack of domain-specific adjudication logic in standard PRMs through the following two-stage process.

\paragraph{Phase I: Offline Verification Principle Distillation.} To enable the model to judge like an expert, we first construct a \textbf{Synthesized Verification Principle Repository} ($\mathcal{P}$) containing generalized adjudication heuristics. Instead of relying on manual engineering or raw trajectory retrieval, we automate the construction process through a \textit{Generate-Contrast-Abstract} pipeline:

\begin{itemize}[leftmargin=*]
    \item \textbf{Contrastive Trajectory Sampling:} Starting from trajectories produced by a GUI agent on an out-of-distribution benchmark, we construct balanced sets of successful and failed episodes. From these episodes, we further sample diverse intermediate steps to cover different process-level supervision scenarios. This procedure emphasizes \textbf{hard negatives}, where the failed action is plausible but logically flawed rather than random noise.
    \item \textbf{Meta-Judge Abstraction:} A strong LLM serves as the Meta-Judge for abstracting verification logic from each contrastive set. Rather than describing only the surface error in a specific interface, the Meta-Judge extracts reusable principles that explain why a transition should be accepted or rejected. For instance, a failed submission with missing required fields can be abstracted into a \textit{State Dependency Validation} principle, where submission actions are valid only after all mandatory fields have been completed.
    \item \textbf{Repository Consolidation:} The extracted logic is consolidated through semantic clustering to keep the repository compact and generalizable. Redundant entries are merged, and overly specific descriptions are rewritten into reusable verification principles. The resulting repository $\mathcal{P}$ covers core GUI verification logic, including state dependencies, functional effectiveness, task-flow compliance, and system feedback validity.
\end{itemize}

\paragraph{Phase II: Online Principle Instantiation.}
At inference time, the PRM converts the principle repository into state-specific criteria for the current decision. Given the task instruction $I$, UI state $S_t$, and a candidate action $(u,a)$, a dense retriever selects the top-$k$ principles from $\mathcal{P}$ that are most relevant to the current intent. The model then instantiates these principles into concrete verification criteria $C_t$, specifying what evidence should be checked in the current UI. For example, a retrieved principle on \textit{State Dependency Validation} becomes a requirement to verify that mandatory fields such as Name and Phone are filled before accepting a \texttt{Submit} action at $(x,y)$. The resulting criteria $C_t$ are passed to Criterion-Guided Autoregressive Perception, where they guide tool selection and evidence gathering.

\subsection{Criterion-Guided Autoregressive Perception}
\label{sec:autoregressive_perception}

Criterion-Guided Autoregressive Perception turns the criteria $C_t$ into a sequential evidence-gathering process. Guided by the instantiated criteria, the model selects visual tools autoregressively and updates the evidence history until the information needed for judgment is sufficient.

\paragraph{Multi-Granularity Visual Toolset.}
To enable multi-scale perception, we define a specialized toolset $\mathcal{T}$ comprising \texttt{OmniParser} for global layout, \texttt{ZoomInSubfigure} for local inspection, \texttt{Point} for spatial grounding, and \texttt{Terminate} for process control (see Appendix~\ref{sec:tool_specs} for technical specifications).

\paragraph{The Autoregressive Investigation Loop.}
Within this perception module, verification is carried out as a dynamic inference trajectory that iteratively refines the evidence available for judgment. At each step $k$, the model conditions on the instruction $I$, the criteria $C_t$, and the evolving observation history $H^{inv}_k$. The loop follows a \textit{Reason-Act-Update} structure:

\begin{itemize}
    \item \textbf{Gap Identification:} 
    The model first maps each criterion in $C_t$ to concrete \textbf{Verification Targets} in the current UI. It then identifies which evidence is still missing for validating these targets. For instance, if a criterion requires checking a toggle state, the model treats the target toggle as unresolved until its precise visual state has been inspected.
    
    \item \textbf{Action Generation:} 
    To bridge the identified gap, the model autoregressively generates the optimal tool action $A_k \in \mathcal{T}$. The selection is strategic: checking existence triggers \texttt{Point}, while resolving visual ambiguity triggers \texttt{ZoomInSubfigure} to isolate the target region. This step represents an active decision to acquire the most discriminative information required for the verdict.
    
    \item \textbf{State Update:} 
    The execution result $O_k$ (e.g., a zoomed local view, point coordinates, or parsed UI structure) is appended to the context, forming the updated history $H^{inv}_{k+1} = H^{inv}_k \cup \{A_k, O_k\}$. This updated state provides a grounded foundation for the next iteration.
\end{itemize}

This loop continues until the model invokes \texttt{Terminate} or the maximum number of iterations is reached, ensuring that the final reward signal is not a hallucinated guess, but a logical conclusion derived from a \textbf{traceable chain of grounded evidence}.

\subsection{Enhanced Supervision with \method}
\label{sec:enhanced_supervision}

In the final stage, \method~aggregates the synthesized knowledge and collected visual facts to produce the supervisory signal. Unlike the standard PRM (Eq.~\ref{eq:prm}) which relies on opaque probability estimation, we employ an \textbf{Evidence-Aware Adjudication Protocol}.

For each candidate $(u, a) \in \mathcal{A}_t$, the Judge Agent rigorously evaluates the alignment between the collected evidence chain $H^{inv}_{\text{final}}$ and the explicit criteria $C_t$. A high reward is assigned only when the evidence affirmatively supports the verification targets.
The final action selection is formalized as:

\begin{equation}
\label{eq:pra}
(u_t, a_t) = \underset{(u, a) \in \mathcal{A}_t}{\text{argmax}} \;\; \mathcal{F}_{\text{judge}}(I, C_t, H^{inv}_{\text{final}}, u, a)
\end{equation}

where $\mathcal{F}_{\text{judge}}$ denotes the scoring function that quantifies the logical consistency between the criteria $C_t$ and the grounded evidence $H^{inv}_{\text{final}}$.

\begin{table*}[t]
\centering

\setlength{\tabcolsep}{12pt} 
\resizebox{\linewidth}{!}{%
\begin{tabular}{@{}l ccc cc cc@{}}
\toprule
\multirow{2}{*}{\textbf{Setting}} & 
\multicolumn{3}{c}{\textbf{DSR (\%)}} & 
\multirow{2}{*}{\textbf{SR (\%)}} & 
\multirow{2}{*}{\textbf{$\Delta$}@1} &
\multirow{2}{*}{\textbf{$\Delta$}@2} \\
\cmidrule(lr){2-4} 
& \textbf{easy} & \textbf{medium} & \textbf{hard} & & & \\
\midrule
\multicolumn{7}{c}{\textbf{\textit{AndroidWorld}}} \\ 
\midrule

InternVL3-8B-Instruct (M3A) & 5.17 & 0.00 & 0.86 & 6.03 & -- & --\\
\quad w/ \textsc{PRM} & 6.90 & 0.00 & \textbf{1.72} & 8.62 & \textcolor{mygreen}{+2.59} & --\\
\rowcolor{mygray} 
\quad w/ \textbf{\textsc{\method}} & \textbf{8.62} & 0.00 & 0.86 & \textbf{9.48} & \textcolor{mygreen}{+3.45} & \textcolor{mygreen}{+0.86}\\
\midrule

InternVL3.5-8B-Instruct (M3A) & 6.04 & 0.00 & 0.86 & 6.90 & -- & --\\
\quad w/ \textsc{PRM} & 9.91 & 0.43 & 0.86 & 11.20 & \textcolor{mygreen}{+4.30} & --\\
\rowcolor{mygray}
\quad w/ \textbf{\textsc{\method}} & \textbf{10.34} & \textbf{1.72} & 0.86 & \textbf{12.93} & \textcolor{mygreen}{+6.03} & \textcolor{mygreen}{+1.73}\\
\midrule

Qwen2.5-VL-7B-Instruct & 15.52 &  3.88 & 0.86 & 20.26 & -- & -- \\
\quad w/ \textsc{PRM} & 17.24 & 1.72 & 0.86 & 19.83 & \textcolor{myred}{-0.43} & -- \\
\rowcolor{mygray}
\quad w/ \textbf{\textsc{\method}} & \textbf{21.55} & \textbf{2.16} & 0.86 & \textbf{24.57} & \textcolor{mygreen}{+4.31} & \textcolor{mygreen}{+4.74} \\
\midrule

Qwen3-VL-8B-Instruct & -- &  -- & -- & 47.60 & -- & -- \\
\quad w/ \textsc{PRM} & 30.60 & 14.66 & \textbf{4.31} & 49.57 & \textcolor{mygreen}{+1.97} & -- \\
\rowcolor{mygray}
\quad w/ \textbf{\textsc{\method}} & \textbf{36.21} & \textbf{14.66} & 3.88 & \textbf{54.74} & \textcolor{mygreen}{+7.14} & \textcolor{mygreen}{+5.17} \\
\midrule

\multicolumn{7}{c}{\textbf{\textit{Mobile-MiniWoB++}}} \\ 
\midrule

Qwen2.5-VL-7B-Instruct & -- & -- & -- & 51.09 & -- & --  \\
\quad w/ \textsc{PRM} & -- & -- & -- & 45.65 & \textcolor{myred}{-5.44} & -- \\
\rowcolor{mygray}
\quad w/ \textbf{\textsc{\method}} & -- & -- & -- & \textbf{53.26} & \textcolor{mygreen}{+2.17} & \textcolor{mygreen}{+7.61}\\
\midrule

Qwen3-VL-8B-Instruct & -- & -- & -- & 59.78 & -- & --  \\
\quad w/ \textsc{PRM} & -- & -- & -- & 59.78 & \textcolor{myred}{0.00} & -- \\
\rowcolor{mygray}
\quad w/ \textbf{\textsc{\method}} & -- & -- & -- & \textbf{65.22} & \textcolor{mygreen}{+5.44} & \textcolor{mygreen}{+5.44}\\
\bottomrule
\end{tabular}%
}
\caption{
    Online execution performance on AndroidWorld and Mobile-MiniWoB++. Each backbone is used as both the execution agent and the reward-model backbone. \textbf{SR} denotes the overall Success Rate, and \textbf{DSR} reports AndroidWorld Success Rate stratified by task difficulty.
    \textbf{$\Delta$@1} is the SR-point gain over the base model, and \textbf{$\Delta$@2} is the SR-point gain of \method~over the Standard PRM.
}
\label{tab:main_result}
\end{table*}

\section{Experiment}

\subsection{Experimental Setup}
\textbf{Benchmarks}. We evaluate on AndroidWorld, Mobile-MiniWoB++, and the mobile subset of OS-Critic Bench, using the standard M3A and Pure Vision execution settings for the online benchmarks. Benchmark-specific details are given in Appendix~\ref{app:setup_details}.

\textbf{Baselines}. We compare \method~with the \textbf{Base Model}, which executes tasks without process supervision, and the \textbf{Standard PRM}, which passively scores candidate actions without explicit criteria or active tool use.

\textbf{Metrics.} We report \textbf{Success Rate (SR)} and \textbf{Difficulty-stratified Success Rate (DSR)}, where Easy, Medium, and Hard DSR values sum to the overall SR.\footnote{Values may not sum exactly because each entry is rounded independently.}

\textbf{Implementation Details}. Online evaluation uses Best-of-$N$ rejection sampling with temperature $\tau=1.0$, where the supervisor selects the highest-scoring candidate step. Offline evaluation directly ranks pre-collected trajectories. Further protocol, system, and repository details are provided in Appendix~\ref{app:setup_details} and Appendix~\ref{app:appendix_data}.

\subsection{Main Results} 

\begin{figure*}[!t]
    \centering
    \includegraphics[width=1.0\linewidth]{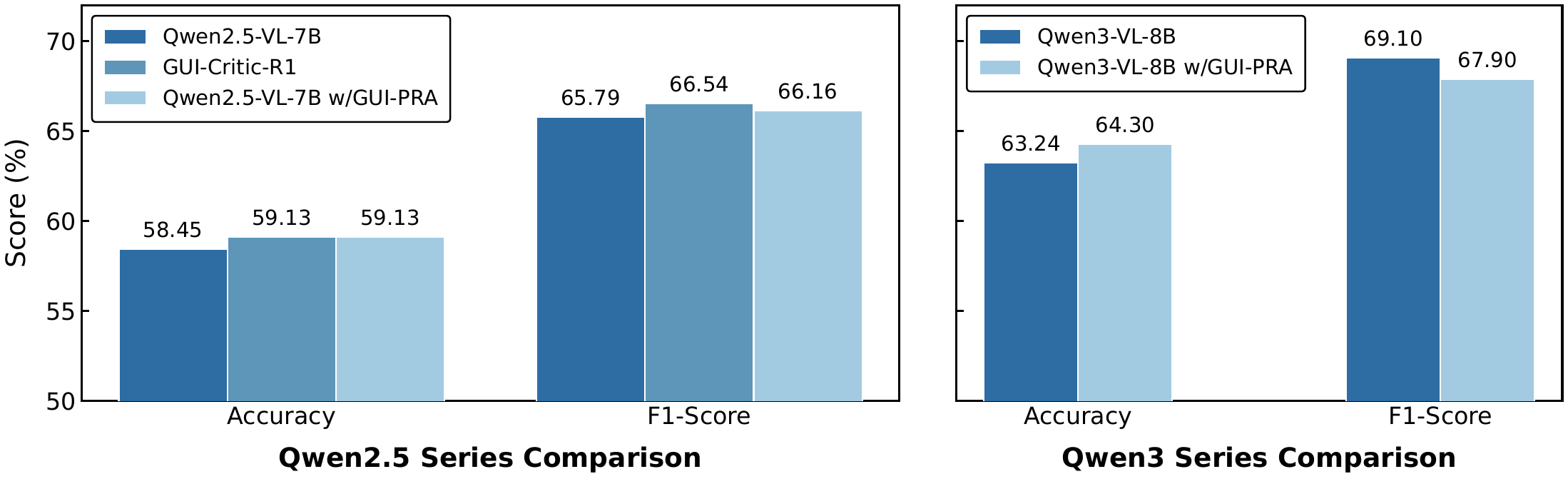}
    \caption{Offline critic performance on the \textbf{OS-Critic Bench} (Mobile Subset). We report Accuracy and F1-Score for zero-shot PRMs, \method, and the supervised GUI-Critic-R1 reference. All Qwen backbone names denote Instruct variants.}
    \label{fig:offline_critic}
\end{figure*}

Table \ref{tab:main_result} and Figure \ref{fig:offline_critic} summarize the performance for online execution and offline discrimination, respectively. Across all settings, \method~consistently outperforms both the autonomous Base Model and the Standard PRM baseline.

\subsubsection{Online Evaluation}

Table~\ref{tab:main_result} reports online execution results on AndroidWorld and Mobile-MiniWoB++. \method~improves over the Standard PRM across all tested backbones and both benchmarks. The strongest result in the main single-backbone comparison comes from Qwen3-VL-8B-Instruct, where \method~raises SR to \textbf{54.74\%} and improves over the Standard PRM by 5.17 SR points. More importantly, the improvement is not limited to already stable settings. On Qwen2.5-VL, the Standard PRM slightly degrades AndroidWorld performance, while \method~recovers the loss and improves over the PRM baseline. Mobile-MiniWoB++ shows the same recovery pattern under a denser UI setting. These results indicate that GUI process supervision benefits from criterion-guided active investigation rather than a passive VLM critic alone.

The Difficulty-Stratified Success Rate further clarifies the source of the gains. Improvements are concentrated in the \textit{Easy} setting and selected \textit{Medium} cases, while \textit{Hard} tasks remain largely unchanged. For Qwen3-VL-8B-Instruct on AndroidWorld, Easy-task DSR rises from 30.60\% to \textbf{36.21\%}, whereas Hard-task DSR stays nearly flat. This pattern matches the verifier role of \method. Active investigation is most effective when the actor has already generated plausible candidate actions; when the correct action is absent from the candidate set, a verifier has limited room to recover. A task-level actor-coverage analysis is provided in Appendix~\ref{app:actor_coverage}.

\subsubsection{Offline Evaluation}
To assess the intrinsic supervision quality of our method, we conduct a static evaluation on the Mobile Subset of the \textbf{OS-Critic Bench}. Figure~\ref{fig:offline_critic} compares \method~with standard zero-shot PRMs and uses \textbf{GUI-Critic-R1} \citep{wanyan2025lookleapguicriticr1model} as a supervised critic reference for the Qwen2.5 series.

The offline results show that active investigation improves transition discrimination without task-specific critic training. On the Qwen2.5 backbone, \method~improves Accuracy from 58.45\% to \textbf{59.13\%} and F1 from 65.79\% to \textbf{66.16\%}. On Qwen3, \method~also raises Accuracy, from 63.24\% to \textbf{64.30\%}, although F1 decreases after the updated official evaluation. We therefore treat the offline result as evidence of improved accuracy and competitive discrimination, rather than a uniform gain on every metric.

\method~also remains competitive with the supervised critic reference. On Qwen2.5, it matches GUI-Critic-R1 in Accuracy and approaches its F1 score. This result complements the online evaluation. Online SR measures whether supervision improves execution, whereas OS-Critic Bench isolates the critic's ability to distinguish valid from invalid transitions. The offline results support our central claim that criterion-guided active investigation can provide stronger evidence for step-level GUI verification than passive scoring.

\subsubsection{Transfer to a Process-Supervised Policy}
\label{sec:process_supervised_policy}
We further evaluate whether \method~transfers to a policy that has already received process supervision. We apply the same frozen-supervisor procedure to the released Perceval-3B-policy checkpoint~\citep{min2026perceval}, a Qwen2.5-VL-3B policy trained with perception-centric process supervision. On the OS-Critic Bench Mobile Subset (Table~\ref{tab:perceval}), \method~improves Accuracy by 0.46 points, Recall by 18.10 points, and F1 by 7.09 points. These results show that \method~reduces false negatives and helps the model make more reliable transition judgments, with a modest precision trade-off.

\begin{table}[t]
\centering
\small
\renewcommand{\arraystretch}{0.9}
\setlength{\tabcolsep}{3.5pt}
\begin{tabular}{@{}l S[table-format=2.2] S[table-format=2.2] S[table-format=2.2] S[table-format=2.2]@{}}
\toprule
\textbf{Configuration} & {\textbf{Acc (\%)}} & {\textbf{Prec (\%)}} & {\textbf{Rec (\%)}} & {\textbf{F1 (\%)}} \\
\midrule
Perceval-3B-policy & 55.48 & 55.96 & 55.20 & 55.58 \\
\addlinespace[2pt]
\rowcolor{mygray}
\quad +\method~ & \textbf{55.94} & 54.73 & \textbf{73.30} & \textbf{62.67} \\
\midrule
\textbf{$\Delta$} & \textcolor{mygreen}{+0.46} & \textcolor{myred}{-1.23} & \textcolor{mygreen}{+18.10} & \textcolor{mygreen}{+7.09} \\
\bottomrule
\end{tabular}
\caption{Perceval-3B policy transfer on the OS-Critic Mobile Subset. Values are percentages.}
\label{tab:perceval}
\end{table}

\subsection{Ablation Study}

\begin{table}
\centering

\small
\renewcommand{\arraystretch}{0.85}
\setlength{\tabcolsep}{4pt}

\begin{tabular}{@{}l S[table-format=2.2] S[table-format=2.2] S[table-format=1.2] S[table-format=2.2] c@{}}
\toprule
\multirow{2}{*}{\textbf{Method}} & 
\multicolumn{3}{c}{\textbf{DSR (\%)}} & 
\multicolumn{1}{c}{\multirow{2}{*}{\textbf{SR (\%)}}} &
\multirow{2}{*}{\textbf{$\Delta$}@1} \\
\cmidrule(lr){2-4}
& {\textbf{easy}} & {\textbf{medium}} & {\textbf{hard}} & & \\
\midrule

PRM & 30.60 & 14.66 & 4.31 & 49.57 & -- \\
\midrule
\addlinespace[2pt]

\rowcolor{mygray}
\textbf{\method~(Full)} & \textbf{36.21} & \textbf{14.66} & \textbf{3.88} & \textbf{54.74} & -- \\

\addlinespace[3pt]
\multicolumn{6}{l}{\textit{w/o component:}} \\ 

\hspace{1em} -- Principle & 36.21 & 12.93 & 2.59 & 51.72 & \textcolor{myred}{-3.02} \\
\hspace{1em} -- Tool   & 32.76 & 12.93 & 4.31 & 50.00 & \textcolor{myred}{-4.74} \\

\bottomrule
\end{tabular}
\caption{
    Ablation study on AndroidWorld with Qwen3-VL-8B-Instruct. \textbf{$\Delta$}@1 reports the SR-point change relative to the full \method~framework.
}
\label{tab:ablation_study}
\end{table}

Table~\ref{tab:ablation_study} ablates the two core modules of \method~on AndroidWorld with Qwen3-VL-8B-Instruct. Removing the active perception tools causes the largest drop, reducing SR by 4.74 points and bringing performance close to the Standard PRM. This result shows that retrieved criteria alone are insufficient when the verifier cannot actively inspect task-relevant UI evidence.

Removing the principle repository also degrades performance, reducing SR by 3.02 points. In this setting, the verifier can still inspect the interface, but lacks explicit adjudication logic for deciding what evidence matters. The two modules therefore address complementary failure sources: principles specify what should be verified, while active perception gathers the evidence needed to verify it.

\subsection{Efficient Supervision and Role Decoupling}

\begin{table}[t]
\centering
\resizebox{0.99\linewidth}{!}{%
\begin{tabular}{@{}lccc@{}}
\toprule
\textbf{Setup} & \textbf{Role} & \textbf{SR (\%)} & \textbf{Impact ($\Delta$)} \\
\midrule
Qwen3-VL-32B-Instruct & Agent Only & 55.60 & -- \\
\quad w/ PRM (8B) & Supervisor & 56.03 & \textcolor{mygreen}{+0.43} \\
\rowcolor{mygray}
\quad w/ \textbf{\textsc{\method}} (8B) & Supervisor & \textbf{57.76} & \textcolor{mygreen}{\textbf{+2.16}} \\
\midrule
Qwen3-VL-8B-Instruct & Agent Only & 47.60 & -- \\
\rowcolor{red!5}
\quad w/ Principle \& Tool  & Integrated Agent & 32.76 & \textcolor{myred}{\textbf{-14.84}} \\
\bottomrule
\end{tabular}%
}
\caption{Efficient supervision and role decoupling on AndroidWorld. \textbf{Impact ($\Delta$)} reports the SR-point change relative to the corresponding Agent Only baseline.}
\label{tab:ablation_scalability}
\end{table}

We further investigate two critical questions:
\begin{rqbox}
    \noindent\textbf{RQ1:} Can a smaller \method~supervisor effectively guide a larger, more capable agent?
    
    \vspace{4pt}
    
    \noindent\textbf{RQ2:} Why should \method~operate as a critic rather than as an executor?
\end{rqbox}

\noindent\textbf{Weak-to-Strong Generalization.} 
For \textbf{RQ1}, we use Qwen3-VL-8B-Instruct as the supervisor for a Qwen3-VL-32B-Instruct agent. As shown in Table~\ref{tab:ablation_scalability}, the standard 8B PRM gives only a marginal gain over the 32B agent. In contrast, the 8B \method~supervisor improves SR by 2.16 points. This result suggests that active verification can make a smaller model useful as a supervisor, since the supervision task focuses on checking a finite set of candidate actions rather than generating the full trajectory.

\noindent\textbf{Critic vs. Actor Paradigm.} 
For \textbf{RQ2}, we test whether the same principle-and-tool mechanism can be integrated directly into the executor. This actor-style deployment performs substantially worse, dropping SR from 47.60\% to 32.76\%. The result supports keeping \method~as a critic. As a decoupled supervisor, \method~can investigate candidate actions and sharpen decision boundaries without mixing verification steps into the actor's generative context.

\subsection{Principle Quality Analysis}
\label{sec:principle_quality}
\begin{table}[t]
\centering
\small
\renewcommand{\arraystretch}{0.9}
\setlength{\tabcolsep}{4pt}
\begin{tabular}{@{}l S[table-format=2.2] S[table-format=2.2] c@{}}
\toprule
\textbf{Configuration} & {\textbf{Acc (\%)}} & {\textbf{F1 (\%)}} & \textbf{$\Delta$Acc} \\
\midrule
Qwen3-VL-8B (Zero-shot PRM)        & 63.24 & 69.10 & -- \\
\addlinespace[2pt]
\quad + \method~(GPT-5-mini)       & 61.64 & 66.13 & \textcolor{myred}{-1.60} \\
\rowcolor{mygray}
\quad + \method~(GPT-5.1)          & \textbf{64.30} & 67.90 & \textcolor{mygreen}{+1.06} \\
\bottomrule
\end{tabular}
\caption{
Principle quality ablation on the OS-Critic Bench (Mobile Subset).
}
\label{tab:principle_quality}
\end{table}

This experiment evaluates whether the criterion module depends on the quality of distilled principles. We keep the runtime backbone fixed as Qwen3-VL-8B-Instruct and vary only the Meta-Judge used for offline distillation. As shown in Table~\ref{tab:principle_quality}, GPT-5-mini principles reduce both Acc and F1 relative to the zero-shot PRM, while GPT-5.1 principles improve Acc by 1.06 points.

The gap comes from the specificity of the distilled verification logic. GPT-5-mini often produces shallow location rules, such as checking whether a tap falls inside a visible bounding box. These rules repeat surface-level grounding cues and can distract the verifier from functional correctness. GPT-5.1 more often distills task-aware rules. For checkout-related goals, it asks the verifier to check whether the selected control belongs to the cart or checkout flow. For termination decisions, it warns against stopping merely because a header or main title is visible. The F1 drop in Table~\ref{tab:principle_quality} also shows a limitation: principle injection is not automatically beneficial, and noisy criteria can weaken downstream supervision.

\subsection{Compute-Budget Comparison}
\label{sec:compute_budget}
\begin{figure}[t]
\centering
\includegraphics[width=0.92\linewidth]{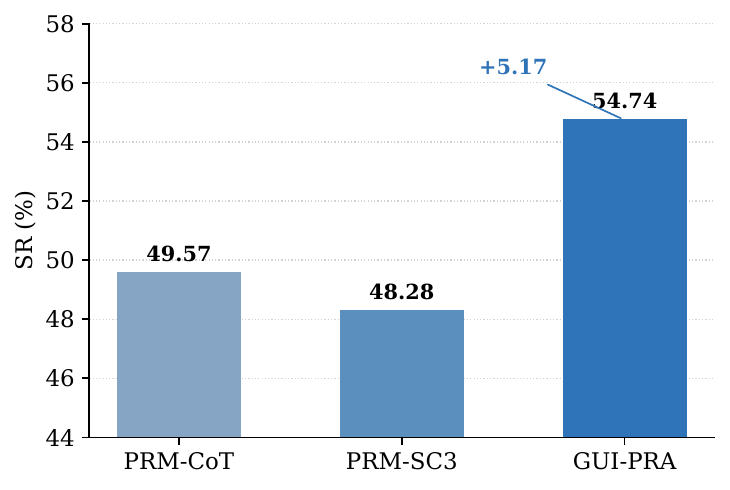}
\caption{
Compute-budget controlled comparison on AndroidWorld (Qwen3-VL-8B-Instruct).
}
\label{fig:compute_budget}
\end{figure}

A natural concern is whether \method's gain merely reflects more test-time compute. We control for this by giving the Standard PRM a matched budget. \textsc{PRM-CoT} uses one chain-of-thought pass per candidate. \textsc{PRM-SC3} is a self-consistency variant that samples each candidate three times at temperature 0.7 and averages the scores. \textsc{GUI-PRA} denotes the full framework.

As shown in Figure~\ref{fig:compute_budget}, self-consistency does not improve the Standard PRM beyond its CoT baseline (48.28 vs.\ 49.57), and still remains more than 6 SR points below \method~(54.74). The result suggests that additional reasoning samples cannot substitute for criterion-guided evidence gathering. The bottleneck is not only how many times the verifier reasons, but whether it knows what to verify and can actively collect the required UI evidence.

\noindent\textbf{Latency is bounded and parallelizable.} \method~adds 6.70 s per step on average with $\approx$1.17 active visual-tool calls per step (1.48 total calls when \texttt{Terminate} is included) relative to the single-pass Standard PRM. Since the $N$ candidate evaluations in each Best-of-$N$ step are independent, the wall-clock overhead can be reduced through parallel execution, although the underlying computation remains. Detailed latency, tool-reliability, and recovery analyses are provided in Appendix~\ref{app:deployment_analysis}.

\subsection{Additional Analyses}
\label{sec:additional_diagnostics}
To further assess the scope and practical behavior of \method, we conduct
additional experiments on actor coverage, latency, failure modes, tool
reliability, and criterion diversity. The results show that \method~benefits
tasks with viable actor candidates, incurs bounded investigation overhead, and
can recover from explicit tool failures, while silent semantic misgrounding
remains a limitation. Detailed results are provided in
Appendices~\ref{app:actor_coverage}, \ref{app:deployment_analysis},
\ref{app:causal_analysis}, and \ref{app:investigation_profile}.

\subsection{Case Study}
Figures~\ref{fig:case_study} and~\ref{fig:case_study_new} present case studies for \textit{ContactsNewContactDraft} and \textit{NotesTodoItemCount}, exemplifying a “do not save” negative constraint and a counting‑completeness requirement respectively. In both cases, a passive verifier can mistake visually plausible progress for task completion or accept premature termination. \method~instead keeps checking task-specific criteria---field population and the save-prohibition rule for the first task, and item-by-item counting completeness for the second---before adjudicating termination. These cases show how criterion-guided supervision enforces functional correctness beyond surface-level visual completion. Detailed operational steps and intermediate reasoning are provided in Appendix~\ref{app:case_study}.

\section{Conclusion}
We introduced \method, a Process Reward Agent that transforms GUI process evaluation from passive scoring into active investigation. By combining Experience-Injected Criterion Synthesis with Criterion-Guided Autoregressive Perception, \method~derives task-specific criteria and gathers grounded UI evidence before assigning supervision. Experiments across online execution and offline critic evaluation show consistent gains over standard PRMs, supporting active verification as a practical direction for reliable GUI agents.

\section*{Limitations}
\method~has shown robust gains across multiple backbones and benchmarks; we note two scope considerations rather than open failure modes.
First, like any MLLM-based verifier, the quality of process supervision is upper-bounded by the underlying visual tools and language backbone, although our experiments suggest that the autoregressive investigation loop already absorbs most of the noise from individual tool calls.
Second, the Principle Repository is currently distilled offline from contrastive trajectories, which already generalizes well across mobile and web environments; extending it into an online, continually updated repository that adapts to entirely new UI paradigms is a natural direction for future work.
\section*{Acknowledgments}
This work is supported by the National Natural Science Foundation of China (No. 62402429, U24A20326, and 62441236), the Ningbo Yongjiang Talent Introduction Programme (No. 2023A-397-G), and the Young Elite Scientists Sponsorship Program by CAST (No. 2024QNRC001). The author gratefully acknowledges the support of Zhejiang University Education Foundation Qizhen Scholar Foundation. This work is supported by Xiaomi Inc.’s research program. Conclusions are the authors’ and do not represent official funder or government policies.%
\bibliography{custom}

\clearpage
\appendix

\section{Tool Details}
\label{sec:tool_specs}

This section provides the specific implementation details for the multi-granularity visual toolset $\mathcal{T}$ employed during the \textit{Criterion-Guided Autoregressive Perception} phase. These tools enable the \method~to actively transition from global understanding to local inspection, resolving visual ambiguities in static screenshots. The input/output specifications are summarized in Table~\ref{tab:ui_tools_summary}, with detailed functional descriptions provided below.

\par \textbf{OmniParser: Global UI Perception}. 
We utilize the OmniParser \citep{lu2024omniparserpurevisionbased} to establish a structural understanding of the global GUI state. This tool operates in two stages to convert a raw screenshot into a semantic representation:
(1) \textbf{Optical Character Recognition (OCR):} It first detects and recognizes all textual elements on the screen, extracting their semantic content and bounding box coordinates.
(2) \textbf{Set-of-Mark (SoM) Annotation:} Leveraging the OCR results and icon detection, the module overlays numeric tags or bounding boxes on interacting elements. 
The output serves as the foundational "map" for the agent, providing a structured textual description (e.g., HTML/XML-like format) alongside the visually annotated image, enabling the LLM to ground its reasoning in the global layout.

\par \textbf{ZoomInSubfigure: Local Detail Disambiguation}. 
To address the \textit{Visual Ambiguity} challenge where fine-grained details (e.g., checkbox states, dense tabular data, or small status icons) are lost in the global view, we introduce the \texttt{ZoomInSubfigure} tool. Specifically, we employ \textbf{Qwen3-VL-8B-Instruct} as the vision backbone to identify and interpret visual content within these specific regions.
\textbf{Input:} A target bounding box $[x_{\min}, y_{\min}, x_{\max}, y_{\max}]$ derived from the OmniParser output or the agent's reasoning.
\textbf{Operation:} The tool crops the original, high-resolution screenshot to the specified region to preserve native pixel density. This cropped view is then processed by the \textbf{Qwen3-VL} backbone to resolve local ambiguities.
\textbf{Output:} A high-fidelity sub-image of the Region of Interest (RoI). This allows the model to inspect visual features (e.g., whether a toggle is "On" or "Off") with pixel-level clarity, effectively removing the noise from the surrounding GUI context.

\par \textbf{Point: Local UI Element Grounding}. 
The \texttt{Point} tool is engineered to validate element existence and determine precise interaction coordinates based on semantic intent. We employ Molmo-7B-D-0924 \citep{deitke2024molmopixmoopenweights}, a state-of-the-art open-weights model, as the backbone.
\begin{itemize}
    \item \textbf{Functionality:} It accepts a natural language description (e.g., "The search icon in the top right corner" or specific button text) and outputs the precise normalized coordinates $(x, y)$.
    \item \textbf{Visual Feedback:} To make the result interpretable for the subsequent verification step, we post-process the output by overlaying a highly visible marker (e.g., a red pentagram) at the predicted coordinates on the screenshot. This explicitly visualizes \textit{where} the agent believes the target element is located, allowing the Judge Agent to verify if the location aligns with the criteria (e.g., "Is the pointed element actually the 'Submit' button?").
\end{itemize}

\par \textbf{Terminate: Investigation Convergence}.
This tool does not perform visual processing but serves as a control signal. It is triggered when the agent determines that the accumulated evidence from the history $H^{inv}$ is sufficient to answer all verification criteria in $C_t$ with high confidence, ending the autoregressive loop.

\newcolumntype{L}{>{\raggedright\arraybackslash}X}

\begin{table*}[!t] 
    \centering
    \resizebox{0.95\linewidth}{!}{%
    \begin{tabularx}{\linewidth}{@{} l L L L @{}} 
        \toprule
        \textbf{Tool} & \textbf{Input} & \textbf{Output} & \textbf{Description} \\
        \midrule
        \texttt{OmniParser} & Image & SoM + BBox & Text-driven object detection \\
        \texttt{Point} & Text + Image & Coordinates & Object localization \\
        \texttt{ZoomInSubfigure} & BBox + Image & Sub-image & Detail disambiguation \\
        \texttt{Terminate} & -- & -- & Terminate loop \\
        \bottomrule
    \end{tabularx}
    }
    \caption{Input and Output Interfaces of the multi-granularity visual toolset $\mathcal{T}$ for interface analysis.}
    \label{tab:ui_tools_summary}
\end{table*}

\section{Comprehensive Implementation Details}
\subsection{Efforts for Reproducibility}
To ensure the reproducibility of our experimental results, we meticulously documented and controlled several key parameters and settings throughout the evaluation of \method.

For the underlying GUI Agent, we deployed the model as a server using the vLLM framework, with the deployment random seed initialized to $42$. To ensure deterministic behavior, the inference temperature was set to $0$ for the base model and all other experimental modules. In contrast, for the online Best-of-N (BoN) sampling experiments, the temperature was set to $1.0$ to encourage diversity. Specifically, we employed distinct random seeds (i.e., $[30, 42, 3407, 114514, 256, 64, 1024, 2]$) to generate the eight candidate trajectories during this phase. During the base model testing phase, a random seed of $42$ was used. 

In the configuration of \method's components, we set the number of criterion sampling attempts to $5$, ensuring the selection of the most relevant experiential references. The maximum number of routing attempts for the autoregressive UI tools component was capped at $4$.

\subsection{Benchmark Details and Evaluation Protocol}
\label{app:setup_details}
\textbf{AndroidWorld}. We evaluate on the standard M3A setting, which uses AndroidWorld as a dynamic benchmark for long-horizon tasks in real applications.

\textbf{Mobile-MiniWoB++}. We use the Pure Vision setting for mobile-style single-page tasks with dense UI elements and fine-grained grounding requirements.

\textbf{OS-Critic Bench}. We evaluate only the mobile subset, since the offline critic experiment is designed to test whether a supervisor can distinguish valid transitions from execution errors.

\textbf{Protocol}. Online evaluation uses Best-of-$N$ rejection sampling with temperature $\tau=1.0$. The supervisor scores each candidate step and selects the highest-scoring action for execution. Offline evaluation ranks pre-collected trajectories directly.

\subsection{Data Construction Statistics}
\label{app:appendix_data}
To construct the principle repository ($P$), we employed a rigorous filtering process. For the online experiment repository, we sampled 50 positive and 50 negative trajectories from the validation set using the Qwen3-VL-8B-Instruct. Following synthesis by the GPT-5.1 Meta-Judge and subsequent deduplication, we obtained a compact set of \textbf{64} verification principles. 
For the offline experiments, we scaled the collection to include 150 positive and 150 negative samples. After iterative merging and refinement, this yielded \textbf{143} and \textbf{119} distinct principles for the Qwen3-VL and Qwen2.5-VL based experiments, respectively.

\section{Additional Analyses}
\label{app:additional_analyses}
This section reports additional analyses of actor coverage, latency, tool
reliability, and reproducibility that complement the main experimental results.

\begin{figure*}[t]
    \centering
    \includegraphics[width=0.98\linewidth]{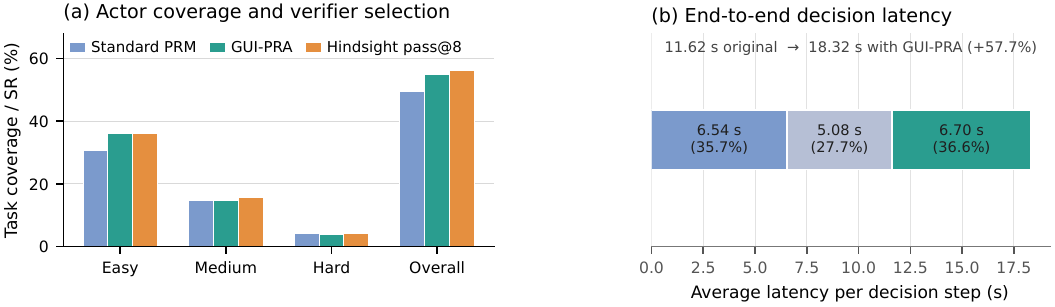}
    \caption{Additional diagnostics. (a) Task-level hindsight coverage measures
    the complete trajectories available to the verifier. (b) The eight-candidate
    decision-time breakdown separates actor generation, passive PRM selection,
    and \method~investigation.}
    \label{fig:additional_diagnostics}
\end{figure*}

\subsection{Actor Coverage and the Hard-Task Boundary}
\label{app:actor_coverage}

\method~is a verifier and selector, not a replacement for long-horizon actor
planning. We therefore ran eight independent Qwen3-VL-8B executions on the same
116 AndroidWorld tasks and computed a hindsight task-level pass@8 reference: a
task is covered when any complete trajectory succeeds. This is an empirical
coverage reference, not a step-level oracle. As shown in Table~\ref{tab:actor_coverage},
the Standard-PRM-to-reference gap is concentrated on Easy tasks (+5.61 points),
with only +0.86 on Medium and no gap on Hard. \method~closes 5.17 of the 6.46
overall points (about 80\%) and matches the Easy reference. Hindsight covers
only 5/116 Hard tasks (4.31\%), so a verifier has little headroom when the actor
has not generated a valid continuation; it can reject a plausible wrong action,
but cannot create a missing branch.

\begin{table}[H]
\centering
\small
\setlength{\tabcolsep}{3.5pt}
\renewcommand{\arraystretch}{0.92}
\caption{Task-level actor coverage on 116 AndroidWorld tasks. Hindsight pass@8 is an empirical complete-trajectory coverage reference, not a step-level oracle.}
\label{tab:actor_coverage}
\begin{tabular}{@{}lrrrr@{}}
\toprule
\textbf{Reference} & \textbf{Easy} & \textbf{Med.} & \textbf{Hard} & \textbf{Overall} \\
\midrule
Standard PRM & 30.60 & 14.66 & 4.31 & 49.57 \\
\rowcolor{mygray}
\textbf{GUI-PRA} & \textbf{36.21} & 14.66 & 3.88 & \textbf{54.74} \\
Hindsight pass@8 & \textbf{36.21} & \textbf{15.52} & \textbf{4.31} & \textbf{56.03} \\
\bottomrule
\end{tabular}
\end{table}

\subsection{Latency, Tool Reliability, and Reproducibility}
\label{app:deployment_analysis}

The diagnostics in this section come from two separately logged AndroidWorld
evaluation snapshots. The latency and tool-call audit uses a 1,475-step runtime
log, whereas the criterion-frequency analysis uses a separate log containing
1,483 distinct evaluation states. These counts refer to different units and
snapshots, so they are not expected to match one-to-one.

The 1,475-step runtime log adds 6.70 seconds per evaluation step and uses
1.17 active visual-tool calls per step, or 1.48 total tool calls when
\texttt{Terminate} is included. In a matched seven-step timing trace, actor
candidate generation ($N=8$), Standard PRM selection, and \method~investigation
take 6.54, 5.08, and 6.70 seconds, respectively, for an estimated 18.32-second
decision. Investigation therefore accounts for 36.6\% of the full decision and
increases the original 11.62-second pipeline by 57.7\% (Figure~\ref{fig:additional_diagnostics}).
Candidate evaluations are independent and can be parallelized, but the
underlying computation remains a substantial cost. A compute control gives
PRM-SC3 48.28\% SR, below Standard PRM CoT (49.57\%) and \method~(54.74\%),
indicating that criterion-relevant evidence, rather than repeated unchanged
scoring, drives the gain. The 6.70-second increment is measured investigation
overhead, not a fixed per-step sequence: calls are made only when evidence is
insufficient and routing is capped at four attempts.

\begin{table}[t]
\centering
\small
\renewcommand{\arraystretch}{1.0}
\setlength{\tabcolsep}{6pt}
\begin{tabular}{@{}llr@{}}
\toprule
\textbf{Metric} & \textbf{Symbol} & \textbf{Value} \\
\midrule
Total evaluation steps        & $N_{\text{step}}$        & 1{,}475 \\
Total wall-clock overhead     & $T_{\text{total}}$       & 9{,}880.0 s \\
Avg.\ overhead per step       & $\bar{T}_{\text{step}}$  & 6.70 s \\
Active visual calls / step & $\bar{C}_{\text{active}}$ & 1.17 \\
Total tool calls / step & $\bar{C}_{\text{total}}$ & 1.48 \\
\bottomrule
\end{tabular}
\caption{
Latency profile of \method~on AndroidWorld. Active visual calls exclude \texttt{Terminate}; total tool calls include it.
}
\label{tab:latency}
\end{table}

\begin{table}[H]
\centering
\small
\setlength{\tabcolsep}{3.5pt}
\renewcommand{\arraystretch}{0.92}
\caption{Timing breakdown under the eight-candidate AndroidWorld protocol.}
\label{tab:latency_breakdown}
\begin{tabular}{@{}lrr@{}}
\toprule
\textbf{Component} & \textbf{s/step} & \textbf{Share} \\
\midrule
Actor generation ($N=8$) & 6.54 & 35.7\% \\
Standard PRM selection & 5.08 & 27.7\% \\
GUI-PRA investigation & 6.70 & 36.6\% \\
\midrule
\textbf{Full GUI-PRA decision} & \textbf{18.32} & \textbf{100.0\%} \\
\bottomrule
\end{tabular}
\end{table}

We audited 164 logged decision steps (Table~\ref{tab:tool_audit}). When OmniParser
or Point explicitly reports missing evidence, the router preserves that gap and
can invoke a complementary tool or choose a state-exposing action; the
Clock/Stopwatch, Wi-Fi/Bluetooth, and Settings traces all recover this way.
Silent semantic misgrounding remains harder: Point repeatedly maps the Google
Lens shortcut to Camera, producing a Home--Lens oscillation and failure.
\method~therefore mitigates overt tool failure but does not guarantee robustness
to syntactically valid, semantically wrong evidence.

\begin{table*}[t]
\centering
\small
\setlength{\tabcolsep}{4pt}
\renewcommand{\arraystretch}{1.05}
\begin{tabularx}{\textwidth}{@{}>{\raggedright\arraybackslash}p{0.22\textwidth} >{\raggedright\arraybackslash}p{0.32\textwidth} X@{}}
\toprule
\textbf{Trace} & \textbf{Tool issue} & \textbf{GUI-PRA response and outcome} \\
\midrule
\texttt{ClockStopWatch}\allowbreak\texttt{Running}\\[-1pt]\textit{Step 0} &
OmniParser returns no relevant element, and Point reports \texttt{Coordinates not found}. &
The router preserves the evidence gap, invokes \texttt{ZoomInSubfigure}, and selects opening the app drawer. The task completes successfully. \\
\texttt{TurnOffWifiAnd}\allowbreak\texttt{TurnOnBluetooth}\\[-1pt]\textit{Step 0} &
No relevant connectivity control is parsed, and Point cannot locate the Wi-Fi icon. &
The router zooms into the status-bar region and selects opening Quick Settings. The task completes successfully. \\
\texttt{TurnOnWifiAnd}\allowbreak\texttt{OpenApp}\\[-1pt]\textit{Step 6} &
OmniParser detects only \texttt{Messages}, and Point cannot locate the Settings app. &
The router invokes \texttt{ZoomInSubfigure} and selects scrolling rather than fabricating a coordinate. Settings becomes visible and is opened successfully. \\
\texttt{CameraTakePhoto}\\[-1pt]\textit{Steps 0/2/4/6/8} &
Point maps the Google Lens shortcut to Camera across repeated steps. &
The apparently valid but incorrect coordinate is repeated, causing Home--Lens oscillation and task failure. \\
\bottomrule
\end{tabularx}
\caption{Tool-reliability audit of 164 logged decision steps. Explicit failures can trigger complementary investigation, whereas silent semantic misgrounding can cascade.}
\label{tab:tool_audit}
\end{table*}

The runtime components are modular and replaceable: OmniParser supplies global
structure, Molmo supplies point grounding, and Qwen3-VL interprets local crops.
Deterministic modules use temperature 0; Best-of-$N$ uses temperature 1.0 with
fixed seeds; criterion synthesis uses five attempts and routing is capped at
four. The implementation, prompts, principle repository, and
repository-construction scripts will be released with the camera-ready
artifacts. Principles are distilled only from AndroidControl \citep{li2024effects}, while AndroidWorld
and web-centric Mobile-MiniWoB++ provide cross-environment evaluation evidence;
desktop and highly dynamic web interfaces remain future work.

\section{Failure-Mode Alignment Analysis}
\label{app:causal_analysis}
\begin{figure}[t]
\centering
\includegraphics[width=0.95\linewidth]{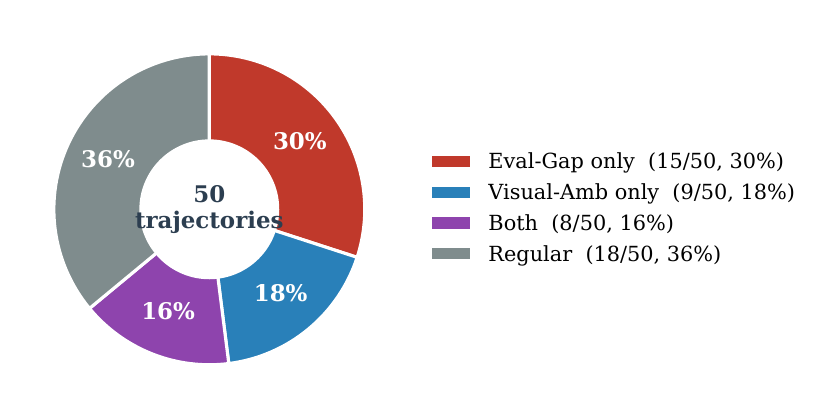}
\caption{Failure-mode breakdown on 50 sampled AndroidWorld trajectories.}
\label{fig:failure_modes}
\end{figure}

We also analyze 50 representative interaction trajectories sampled from AndroidWorld and manually annotate the main reason why the Standard PRM gives an unreliable score. As shown in Figure~\ref{fig:failure_modes}, 46\% of the sampled trajectories involve an evaluative knowledge gap, and 34\% involve visual ambiguity. Some trajectories involve both. The 16\% overlap suggests that the two problems often appear together, which is why we couple criterion synthesis with active perception.

The examples below show what these categories mean in practice. On \textit{SportsTracker\allowbreak Activities\allowbreak Count\allowbreak For\allowbreak Week}, the Standard PRM over-rewards a premature \texttt{terminate} action because the app's main title is already visible. \method~instead applies a criterion requiring the evaluator to confirm that the expected interaction pattern has been completed, and it selects a \texttt{swipe} action that exposes the next task-relevant state. On dense-UI tasks such as \textit{MarkorDeleteNewestNote} and \textit{ExpenseDeleteDuplicates2}, static global-view scoring cannot reliably distinguish visually similar candidates. In these cases, \method~invokes \texttt{ZoomInSubfigure} to isolate the relevant region and ground the verdict in local visual evidence. This analysis complements the aggregate results and helps explain why the gains concentrate on solvable Easy and Medium tasks, while Table~\ref{tab:ablation_study} shows that both principles and tools are necessary.

\section{Investigation Profile and Criterion Diversity}
\label{app:investigation_profile}
\begin{figure*}[t]
\centering
\begin{subfigure}[t]{0.49\linewidth}
\centering
\includegraphics[width=\linewidth]{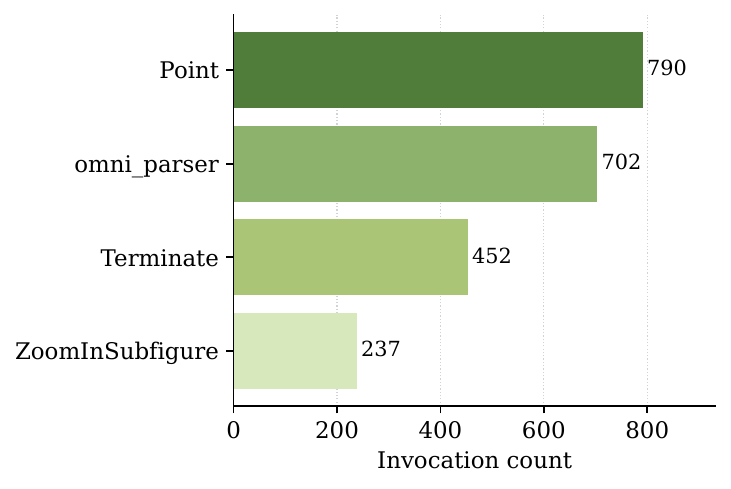}
\subcaption{Tool invocation frequency across 1{,}475 runtime decision steps. \method~makes an average of 1.17 active visual-tool calls per step, or 1.48 total calls when \texttt{Terminate} is included.}
\label{fig:tool_freq}
\end{subfigure}\hfill
\begin{subfigure}[t]{0.49\linewidth}
\centering
\includegraphics[width=\linewidth]{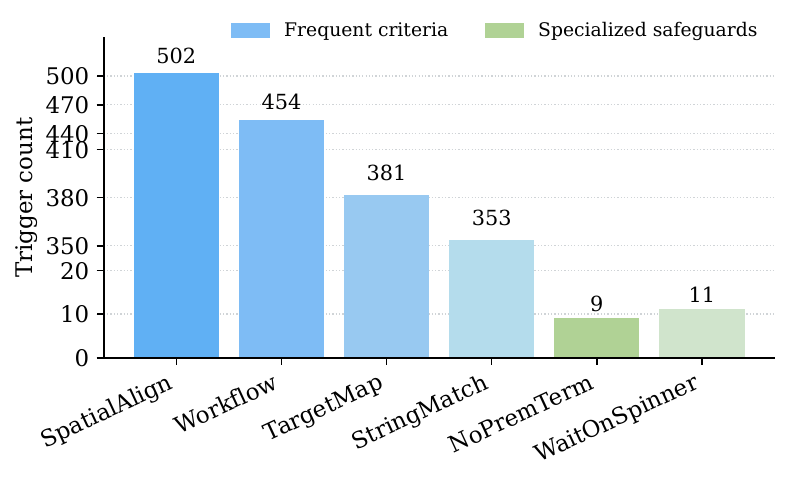}
\subcaption{Criterion utilization with a piecewise scale. In a separate 1{,}483-state criterion-analysis log, \method~instantiates 61 unique criteria, with frequently reused checks and rarer specialized safeguards.}
\label{fig:criterion_freq}
\end{subfigure}
\caption{
\textbf{Investigation profile of \method~on AndroidWorld (Qwen3-VL-8B-Instruct).} The two panels use separately logged snapshots and therefore have different denominators.
(\subref{fig:tool_freq}) Tool calls are selectively allocated across routine grounding and local inspection.
(\subref{fig:criterion_freq}) Criteria combine frequently reused verification routines with specialized safeguards for rarer UI states.
}
\label{fig:investigation_profile}
\end{figure*}

We inspect the investigation process by logging tool calls and synthesized criteria during the AndroidWorld run with Qwen3-VL-8B-Instruct (Figure~\ref{fig:investigation_profile}). The tool-call panel uses the 1,475-step runtime log described above, while the criterion panel uses the separate 1,483-state criterion-analysis log.

\noindent\textbf{Tool calls are selectively allocated.} Across 1{,}475 runtime decision steps, \method~invokes \texttt{Point} 790 times, \texttt{omni\_parser} 702 times, \texttt{ZoomInSubfigure} 237 times, and \texttt{Terminate} 452 times. The first three tools total 1,729 calls, or $\approx$1.17 active visual-tool calls per step; including \texttt{Terminate}, the total is 2,181 calls, or $\approx$1.48 calls per step. \texttt{Point} and \texttt{omni\_parser} account for most active calls because they provide routine spatial grounding and global layout evidence. \texttt{ZoomInSubfigure} is used less frequently, suggesting that local inspection is triggered selectively when global evidence is insufficient rather than being applied uniformly as extra computation.

\noindent\textbf{Criteria remain state-adaptive.} Out of 1{,}483 distinct evaluation states, \method~instantiates 61 unique criteria. Nine head criteria are triggered more than 300 times each, and 22 are triggered more than 100 times, indicating that the model repeatedly uses several broad GUI-logic dimensions rather than collapsing to a single fallback rule. The four most frequent criteria correspond to dialog spatial alignment (502), workflow progression (454), target mapping for lists and icons (381), and exact string matching for text entry (353). Long-tail criteria handle rarer but consequential states. For example, the rule ``\textit{do not terminate solely because a main title or header is visible}'' triggers 9 times, while ``\textit{when a loading indicator is visible, prefer a wait action}'' triggers 11 times. This distribution indicates that the Principle Repository provides both reusable verification routines and specialized safeguards for deceptive UI states.

\section{Detailed Case Study}
\label{app:case_study}

\begin{figure*}[t]
    \centering
    \includegraphics[width=1\linewidth]{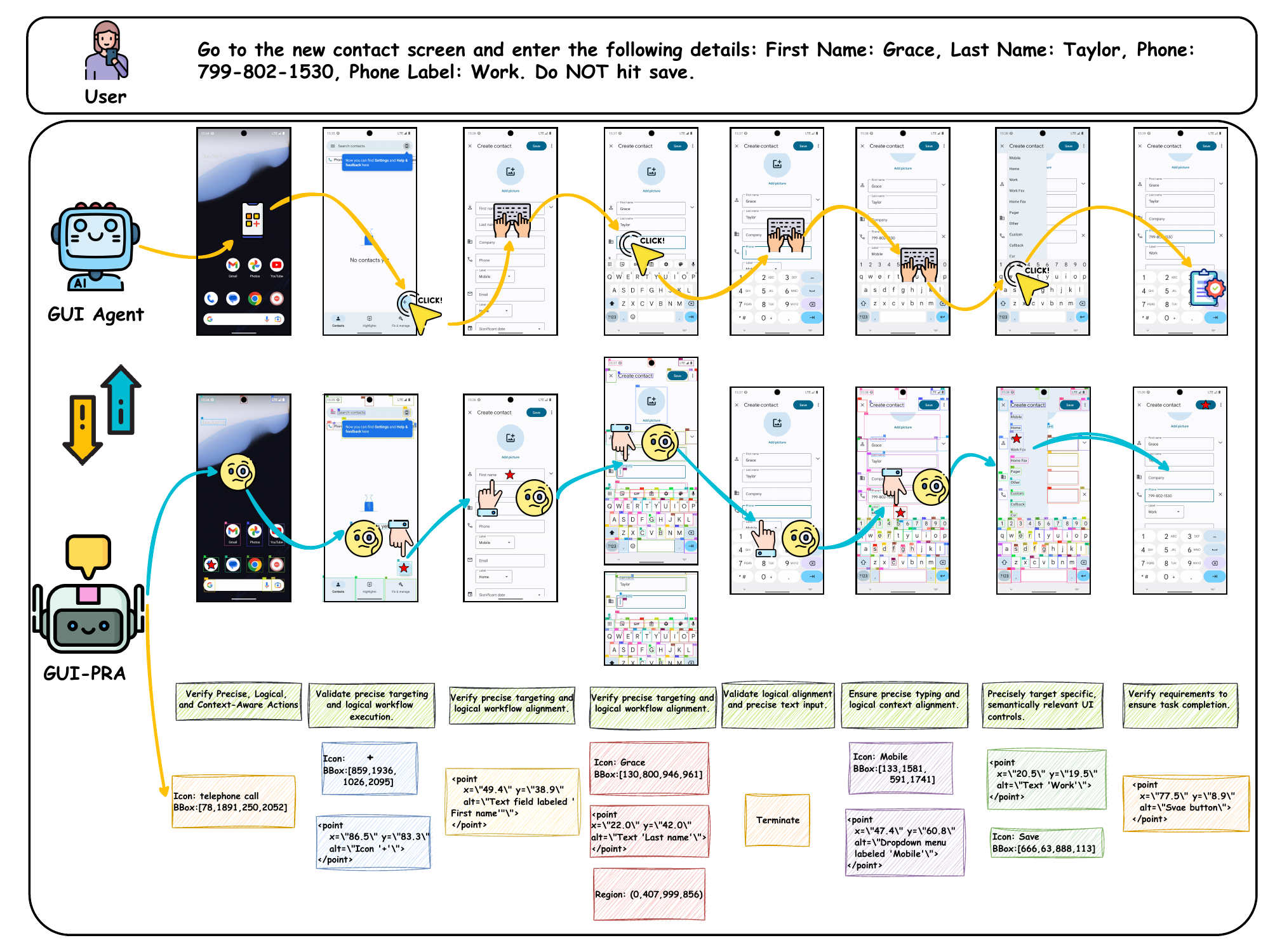}
    \caption{A complete case of \method~guiding a GUI Agent to complete the 'ContactsNewContactDraft' task. The figure illustrates the parallel process flows, showing the agent's action trajectory (top row) and the continuous supervision provided by \method~(bottom row) across multiple steps until task completion.}
    \label{fig:case_study}
    \vspace{-0.5cm}
\end{figure*}

Figure~\ref{fig:case_study} illustrates a complete operational flow of our \method~framework, showcasing its interaction with a base GUI Agent to fulfill a user's request. The top row depicts the trajectory of the base GUI Agent. It correctly executes all the positive data entry sub-tasks: it navigates to the contacts application, initiates the creation of a new contact, and accurately inputs the name, phone number, and label.  

In parallel, the bottom row shows the continuous monitoring and reasoning process of our \method. At each step, \method~leverages its UI Tools to perceive the screen, grounding the agent's actions and the state of the UI elements, as evidenced by the `Icon` and `<point>` outputs. The primary challenge here is not the data entry itself, but correctly interpreting the negative constraint: ``Do NOT hit save.'' \method~excels by continuously validating the agent's progress against this complex goal. It correctly determines the precise moment the task is finished, instructing the agent to terminate rather than incorrectly proceeding to save.

This intervention prevents the GUI Agent from making an irreversible error that would have resulted in task failure. This case highlights \method's ability to provide nuanced, process-level supervision that goes beyond simple action validation, ensuring strict adherence to complex user constraints.

\begin{figure*}[!t]
    \centering
    \includegraphics[width=\textwidth]{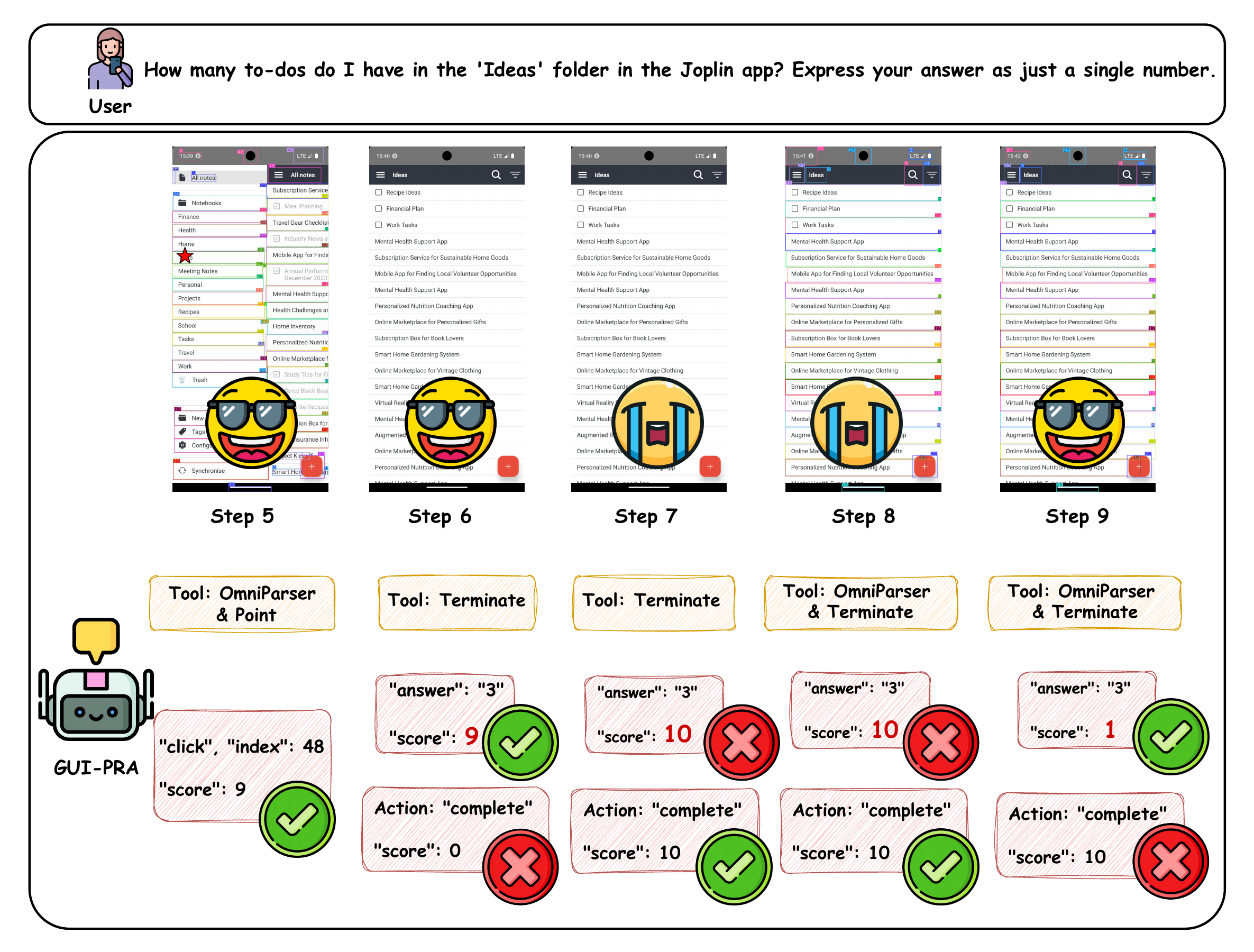}
    \caption{A second case on the \texttt{NotesTodoItemCount} task. The agent prematurely terminates before all todo items have actually been observed, while \method~enforces a counting-completeness criterion and only accepts termination after every relevant item has been verified through active perception.}
    \label{fig:case_study_new}
    \vspace{-0.5cm}
\end{figure*}

As shown in Figure~\ref{fig:case_study_new}, this second example illustrates the same supervision pattern on a different task type. On \texttt{NotesTodoItemCount}, the goal requires the agent to enumerate todo items inside a notes view and report a final count. The base agent tends to terminate based on a partial view of the list, returning a count that omits items below the fold or hidden inside collapsed sections. \method~instead instantiates a counting-completeness criterion that explicitly requires verifying whether every relevant item has been observed, and uses active perception to inspect the list region before adjudicating termination. Together, the two cases show that process supervision is most useful when the verifier must enforce task-specific completion semantics---negative constraints in \texttt{ContactsNewContactDraft} and counting completeness in \texttt{NotesTodoItemCount}---rather than a single generic notion of progress.

\section{PROMPTS}
We provide the prompts in constructing \method~below.

\newtcolorbox{mybox}[1]{
    title={#1},
    colback=white,
    colframe=gray!70,
    fonttitle=\bfseries,
    boxsep=5pt,
    left=5pt,right=5pt,top=5pt,bottom=5pt
}

\newtcolorbox{fullwidthbox}[1]{
    title={#1},
    colback=white,
    colframe=gray!70,
    fonttitle=\bfseries,
    boxsep=5pt,
    left=5pt,right=5pt,top=5pt,bottom=5pt,
    titlerule=0.5pt,
    breakable
}

\subsection{\method: Experience-Injected Criterion Synthesis}

    \begin{fullwidthbox}{\method: Criterion retrieval}
\textbf{\texttt{SYSTEM:}} \\
You are an expert in Android UI automation. Select the 5 most relevant experiences/guidelines for the current situation.
\\ \\
\textbf{\texttt{USER:}} \\
Current Goal: \{goal\}\\
\textcolor{myred} {\textbf{Available Experiences:}}
\{exp\_list\_str\}\\
Select the 5 most relevant experiences that help with the current step.
Output ONLY the indices of the selected experiences as a comma-separated list (e.g., 0, 5, 12).

    \end{fullwidthbox}

\subsection{\method: Criterion-Guided Autoregressive Perception}

\begin{fullwidthbox}{\method: UI Tool Routing}
\textbf{\texttt{SYSTEM:}} \par
You are a visual assistant for GUI process verification. Your goal is to inspect the current UI state and collect only the visual evidence needed by a downstream process-reward judge. Do not answer the task yourself.

\vspace{1em}
\vspace{0.5em}
Here are the available tools:
\begin{itemize}[leftmargin=*, labelindent=0pt]
    \item \textbf{\texttt{Point}}: Identifies a specific point in the image based on a description and returns coordinates. Use when the current evidence does not localize a required UI icon or text target. Example: \par
    \texttt{\{"name": "Point", "arguments": \{"image": "img\_1", "param": "Icon 'Gmail'"\}\}}
    
    \item \textbf{\texttt{omni\_parser}}: Parses a UI or general image to detect, label, and locate all interactive and non-interactive elements like icons, buttons, and text fields. Use it when a global inventory is needed and the current evidence is incomplete. Example: \par
    \texttt{\{"name": "omni\_parser", "arguments": \{"image": "img\_1"\}\}}
    
    \item \textbf{\texttt{ZoomInSubfigure}}: Crops the image to the specified subfigure. Use when a specific region remains visually ambiguous or illegible. Example: \par
    \texttt{\{"name": "ZoomInSubfigure", "arguments": \{"image": "img\_1", "param": "File Creation Panel: Naming \& Format"\}\}}
    
    \item \textbf{\texttt{Terminate}}: Ends the investigation and returns the result for downstream scoring. Use when all required evidence has been gathered. Example: \par
    \texttt{\{"name": "Terminate", "arguments": \{"ans": "1985"\}\}}
\end{itemize}

To gather relevant information:
\begin{enumerate}[leftmargin=*, label=\arabic*.]
    \item Assess the type of question provided. Based on the problem type (e.g., locating specific data points, extracting text, analyzing regions, etc.), select the most suitable tool or combination of tools.
    \item If segmentation or line drawing requires coordinates that are not already available, use the \texttt{Point} tool to identify them.
    \item Use the selected tools logically and sequentially, with each tool building on the output of the previous one.
    \item Use a visual tool only for an evidence gap; if no gap remains, call \texttt{Terminate}.
\end{enumerate}

\vspace{1em}
\textbf{Example Output:}

\textit{Example 1:}
\begin{jsonexample}
\{ \\
\quad "thought": "The current evidence does not expose all interactive elements needed to evaluate the candidate action. I will use the omni\_parser tool to fill this specific evidence gap and provide the downstream judge with a global UI inventory.", \\
\quad "actions": [ \\
\quad\quad \{"name": "omni\_parser", "arguments": \{"image": "img\_1"\}\} \\
\quad ] \\
\}
\end{jsonexample}

\vspace{0.5em}
\textit{Example 2:}
\begin{jsonexample}
\{ \\
\quad "thought": "The current evidence does not localize the text target needed to evaluate the candidate action. I will use the Point tool to obtain the coordinates of 'Sun, Oct 15' and close this evidence gap.", \\
\quad "actions": [ \\
\quad\quad \{"name": "Point", "arguments": \{"image": "img\_1", "param": "Text 'Sun,Oct 15'"\}\} \\
\quad ] \\
\}
\end{jsonexample}

\begin{samepage}
\vspace{0.5em}
\textit{Example 3:}
\nopagebreak[4]
\begin{jsonexample}
\{ \\
\quad "thought": "The current view leaves the file-naming field ambiguous, so the evidence is insufficient for evaluation. I will use ZoomInSubfigure to isolate 'File Creation Panel: Naming \& Format' and inspect that region at higher resolution.", \\
\quad "actions": [ \\
\quad\quad \{"name": "ZoomInSubfigure", "arguments": \{"image": "img\_1", "param": "File Creation Panel: Naming \& Format"\}\} \\
\quad ] \\
\}
\end{jsonexample}
\end{samepage}

If further action is required, continue building on the previous step with the correct tool.
\end{fullwidthbox}

\begin{fullwidthbox}{\method: UI Tool Routing}
\textbf{\texttt{USER:}}
\par
User Question: \promptvar{initial\_prompt} \\
\par
\textcolor{myred} {\textbf{Relevant Judge Experiences/Guidelines}}: \promptvar{experiences\_text}
\par
You have already taken some steps. Here is the history of your actions and their observations:
\par
Current tool calling history: \promptvar{history\_str}

\vspace{1em}
\textbf{Your Task (OI - Observation \& Introspection):}
\par
\textbf{Summarize}: Briefly summarize what you have learned from the history.
\par
\textbf{Decide}: Based on your summary and the initial goal, decide on the next step. Is the evidence sufficient for the downstream process-reward judge to evaluate the current candidate action?

\begin{itemize}[leftmargin=*, labelindent=0pt, topsep=2pt, itemsep=2pt]
    \item \textbf{If YES}, call the tool:
    \texttt{\{"name": "Terminate", "arguments": \{"ans": "<your final answer>"\}\}}
    
    \item \textbf{If NO}, call the tool that directly addresses the identified evidence gap.
\end{itemize}

\textbf{Avoid repeats unless the UI state changes.}

\end{fullwidthbox}

\subsection{\method: Best of N Selection}
\label{app:best_of_N}

\begin{fullwidthbox}{\method: BoN Selection}
\textbf{\texttt{SYSTEM:}} \par
You are a Process Reward Model (PRM). Evaluate the candidate action based on the user's instruction, screen image, and history.\\

\textbf{GUI Agent Action Space and Execution Rules.} A candidate action must follow the declared GUI-agent JSON schema and be executable from the current screen state. Evaluate one action at a time; treat \texttt{Terminate} as valid only when the task is complete and all explicit constraints are satisfied.\\

\vspace{0.5em}
\textbf{CRITICAL: Task Constraint Compliance (MUST CHECK FIRST)} \par
Before scoring, identify ALL explicit constraints in the user's instruction (e.g., "in order", "one by one", "first... then...", "all of", specific sequences).
\begin{itemize}[leftmargin=0.5cm, labelindent=0pt]
    \item If the action \textbf{violates any explicit constraint}, the score MUST be $\le$ 3, regardless of how "reasonable" the action seems otherwise.
    \item Example: If instruction says "add songs in order: A, B, C" and the action adds B first, this is a FATAL violation (score 0-2).
\end{itemize}

\vspace{0.5em}
\textbf{Evaluation Criteria (in priority order):}
\begin{enumerate}[leftmargin=0.5cm, label=\arabic*.]
    \item \textbf{Constraint Compliance}: Does the action respect ALL explicit requirements in the instruction? (e.g., order, sequence, completeness). Violations = automatic low score.
    \item \textbf{Reasoning Accuracy}: Does the reasoning correctly understand the current state AND the task requirements? Penalize reasoning that ignores or misinterprets constraints.
    \item \textbf{Progress}: Does the action move toward the goal in the CORRECT way (respecting constraints)?
    \item \textbf{Efficiency}: Is it optimal? Avoids loops and redundancy.
    \item \textbf{Format Compliance}: Action JSON must follow the defined schema strictly.
\end{enumerate}

\vspace{0.5em}
\textbf{Scoring (0-10) - Be Discriminative:}
\begin{itemize}[leftmargin=0.5cm, labelindent=0pt]
    \item \textbf{9-10: Perfect} - respects ALL constraints, correct reasoning, optimal action.
    \item \textbf{7-8: Good} - respects constraints, minor inefficiencies or slightly suboptimal path.
    \item \textbf{4-6: Acceptable} - respects constraints but has weak reasoning or unnecessary steps.
    \item \textbf{2-3: Poor} - violates a constraint OR has significant reasoning errors.
    \item \textbf{0-1: Critical failure} - violates core task constraints, completely wrong action, or false completion.
\end{itemize}

\vspace{0.5em}
\textbf{Output Format:} \par
First, provide a brief explanation of your evaluation, detailing why you assigned the score. \par
Then, output the score enclosed within \texttt{<score>} and \texttt{</score>} tags.
\end{fullwidthbox}

\begin{fullwidthbox}{\method: BoN Selection}
\textbf{\texttt{USER:}} \par
Please evaluate the following candidate action.

\vspace{0.5em}
\textbf{User's Instruction:}
\begin{itemize}[leftmargin=0.5cm, labelindent=0pt]
    \item \textbf{User Goal:} \{goal\}
    \item \textbf{Action History:} \par \{history\}
    \item \textbf{Relevant Experiences:} \par \{experiences\}
\end{itemize}
Evaluate the candidate action based on the background information provided above. Strictly follow the content and requirements mentioned in the Agent Capabilities and Guidelines. Ensure the action is valid within the current UI state and history.

\vspace{0.5em}
\textbf{Candidate Action to Evaluate:} \par
\{action\}

\vspace{0.5em}
\textbf{Tool Information:} \par
The information below is tool-using information which will help you better score the candidate action in the PRM stage. \\
Relevant tool introductions:
\begin{itemize}[leftmargin=0.5cm, labelindent=0pt]
    \item \textbf{\texttt{Point}}: Identifies a specific point in the image based on a description and returns coordinates. Use when the current evidence does not localize a required UI icon or text target. Then use a red five-pointed star to highlight the coordinates in the image. Example: \texttt{\{"name": "Point", "arguments": \{"image": "img\_1", "param": "Icon 'Gmail'"\}\}}
    
    \item \textbf{\texttt{omni\_parser}}: Parses a UI or general image to detect, label, and locate all interactive and non-interactive elements like icons, buttons, and text fields. Use it when a global inventory is needed and the current evidence is incomplete. Example: \texttt{\{"name": "omni\_parser", "arguments": \{"image": "img\_1"\}\}}
    
    \item \textbf{\texttt{ZoomInSubfigure}}: Crops the image to the specified subfigure. Use when a specific region remains visually ambiguous or illegible. Example: \texttt{\{"name": "ZoomInSubfigure", "arguments": \{"image": "img\_1", "param": "File Creation Panel: Naming \& Format"\}\}}
\end{itemize}

TOOL\_INFORMATION: \{tool\_chain\_filter\}

\vspace{0.5em}
\textit{[Attached Screen Images \& Tool Result Images]}

\vspace{0.5em}
Based on the tool information, carefully analyze the candidate action's correctness and format, and assign a granular score based on how helpful the action is for completing the task.
\end{fullwidthbox}

\subsection{PRM: Best of N Selection}

 \begin{fullwidthbox}{PRM: BoN Selection}
\textbf{\texttt{SYSTEM:}} \par
You are a \textbf{Process Reward Model}. Your task is to evaluate a single \textbf{candidate action step} based on a \textbf{user's instruction} and \textbf{provided screen image} (screenshots of the most recent steps, with a maximum of 5).

\vspace{0.5em}
\textit{\{task\_template\}}

\vspace{0.5em}
\textbf{Evaluation Process:}
\begin{enumerate}[leftmargin=0.5cm, label=\arabic*.]
    \item \textbf{Understand the Goal:} Carefully review the user's instruction and the current screen image.
    \item \textbf{Determine Your Optimal Action:} Based on the instruction and image, decide what you believe is the best possible action step.
    \item \textbf{Evaluate the Candidate Action:} Compare the provided candidate action step against your optimal action.
    \item \textbf{Assign a Score:} Assign a numerical score to the candidate action from 0 to 100. If the candidate action is correct and has a correct reasoning process, a higher score should be given.
\end{enumerate}

\textbf{Output Format:} \par
First, provide a brief explanation of your evaluation, detailing why you assigned the score. \par
Then, output the score enclosed within \texttt{<score>} and \texttt{</score>} tags.
\end{fullwidthbox}

\begin{fullwidthbox}{PRM: BoN Selection}
\textbf{\texttt{USER:}} \par
Please evaluate the following candidate action based on the user's instruction and the provided screen image (screenshots of the most recent steps, with a maximum of 5), following all guidelines from the system prompt.

\vspace{0.5em}
\textbf{User's Instruction:} \par
\{goal\}

\vspace{0.5em}
\textbf{History:} \par
\{history\}

\vspace{0.5em}
\textbf{Candidate Action to Evaluate:} \par
\{candidate\_response\}

\vspace{0.5em}
Your evaluation should include a reasoning process and a score wrapped in \texttt{<score></score>} tags.
\end{fullwidthbox}

\section{Statement on the Usage of Large Language Models}

During the preparation of this manuscript, a Large Language Model (LLM) was utilized as an auxiliary tool. Its application was strictly limited to improving the language and readability of the text, as well as assisting with the formatting of figures. The authors have meticulously reviewed and edited all machine-generated suggestions to ensure the scientific accuracy and integrity of the final content, for which they take full responsibility.

\end{document}